\pdfoutput=1

\documentclass[11pt]{article}

\usepackage[preprint]{acl}

\usepackage{times}
\usepackage{latexsym}

\usepackage[T1]{fontenc}

\usepackage[utf8]{inputenc}

\usepackage{microtype}

\usepackage{inconsolata}

\usepackage{graphicx}

\usepackage[symbol]{footmisc}
\usepackage{lipsum}

\newcommand\blfootnote[1]{%
  \begingroup
  \renewcommand\thefootnote{}\footnote{#1}%
  \addtocounter{footnote}{-1}%
  \endgroup
}

\usepackage[nolist]{acronym}
\begin{acronym}
\acro{nlp}[NLP]{Natural Language Processing}
\acro{pos}[PoS]{Part of Speech}
\acro{plm}[PLM]{Pretrained Language Model}
\acroplural{plm}[PLMs]{Pretrained Language Models}
\end{acronym}

\usepackage{dirtytalk}
\usepackage{listings}
\usepackage{xcolor}
\usepackage{refcount}

\colorlet{punct}{red!60!black}
\definecolor{background}{HTML}{EEEEEE}
\definecolor{delim}{RGB}{20,105,176}
\colorlet{numb}{magenta!60!black}

\lstdefinelanguage{json}{
    basicstyle=\normalfont\ttfamily,
    stepnumber=1,
    showstringspaces=false,
    breaklines=true,
    frame=lines,
    backgroundcolor=\color{background},
    literate=
     *{0}{{{\color{numb}0}}}{1}
      {1}{{{\color{numb}1}}}{1}
      {2}{{{\color{numb}2}}}{1}
      {3}{{{\color{numb}3}}}{1}
      {4}{{{\color{numb}4}}}{1}
      {5}{{{\color{numb}5}}}{1}
      {6}{{{\color{numb}6}}}{1}
      {7}{{{\color{numb}7}}}{1}
      {8}{{{\color{numb}8}}}{1}
      {9}{{{\color{numb}9}}}{1}
      {:}{{{\color{punct}{:}}}}{1}
      {,}{{{\color{punct}{,}}}}{1}
      {\{}{{{\color{delim}{\{}}}}{1}
      {\}}{{{\color{delim}{\}}}}}{1}
      {[}{{{\color{delim}{[}}}}{1}
      {]}{{{\color{delim}{]}}}}{1},
}

\title{An Improved Method for Class-specific Keyword Extraction: \\ A Case Study in the German Business Registry}

\author{Stephen Meisenbacher$^\spadesuit$\textsuperscript{1}, Tim Schopf$^\spadesuit$\textsuperscript{1}, \\ \textbf{Weixin Yan\textsuperscript{1}, Patrick Holl\textsuperscript{2}, and Florian Matthes\textsuperscript{1}} \\
          \textsuperscript{1}Technical University of Munich \\ School of Computation, Information and Technology \\ Department of Computer Science, Garching, Germany \\
            \textsuperscript{2}Fusionbase GmbH, Munich, Germany\\
        \texttt{\{first.last\}@tum.de, patrick.holl@fusionbase.com, matthes@tum.de}}

\begin{document}
\maketitle
\begin{abstract}
The task of \textit{keyword extraction} is often an important initial step in unsupervised information extraction, forming the basis for tasks such as topic modeling or document classification. While recent methods have proven to be quite effective in the extraction of keywords, the identification of \textit{class-specific} keywords, or only those pertaining to a predefined class, remains challenging. In this work, we propose an improved method for class-specific keyword extraction, which builds upon the popular \textsc{KeyBERT} library to identify only keywords related to a class described by \textit{seed keywords}. We test this method using a dataset of German business registry entries, where the goal is to classify each business according to an economic sector. Our results reveal that our method greatly improves upon previous approaches, setting a new standard for \textit{class-specific} keyword extraction.
\end{abstract}

\section{Introduction}
\blfootnote{$^\spadesuit$Equal contribution}
As the amount of information created daily continues to rise in the age of big data \cite{chen2014big}, a core challenge becomes how to extract valuable structured information from largely unstructured text documents \cite{7359270,song-etal-2023-survey}. An important first step in the process of Information Retrieval (IR) is often the extraction of keywords (or phrases) from documents, which can provide an initial clue about the information stored within the document \cite{Firoozeh_Nazarenko_Alizon_Daille_2020,XIE2023103382}. With the extraction of meaningful keywords, NLP tasks such as Topic Modeling or Document Classification can be bootstrapped.

Over the past few decades, a number of unsupervised keyword extraction approaches have been proposed in the literature, ranging from frequency-based methods to statistics-based methods \cite{Firoozeh_Nazarenko_Alizon_Daille_2020}, and more recently, methods using graphs or leveraging the capabilities of transformer-based language models \cite{nomoto2022keyword,tran2023recent}. Supervised approaches have been proposed, with the downside of requiring reliable training data \cite{Firoozeh_Nazarenko_Alizon_Daille_2020}.

While a myriad of keyword extraction approaches has appeared in the literature, they are often of the \textit{unguided} nature, where any relevant keywords are extracted regardless of the downstream goal. As such, there has been a scarcity of research in the direction of \textit{class-specific} keyword extraction, where only keywords adhering to a particular \textit{class} are extracted. Presumably, this type of keyword extraction would be useful in settings where a targeted set of keywords is desired, rather than any relevant keyword in a document.

To address this open research challenge, we devise a novel class-specific keyword extraction pipeline, which builds upon the popular open-source package \textsc{KeyBERT}\footnote{\url{https://maartengr.github.io/KeyBERT/}} \cite{maarten_grootendorst_2023_8388690}. We envision an iterative process which is guided by user-provided \textit{seed keywords}. With these, candidate keywords are ranked according to a two-part scoring scheme, and the seed keywords are augmented by top candidates from each iteration.

We evaluate our approach on a dataset of German business registry (\textit{Handelsregister}) entries, where the goal is to extract as many \textit{class-specific} keywords according to \textit{economic sectors}, as defined by an existing classification scheme. In this evaluation, we show that our method greatly outperforms previous keyword extraction methods, demonstrating the strength of our approach in extracting class-specific keywords.

The contributions of our work are as follows:
\begin{enumerate}
    \itemsep -0.3em
    \item We address the task of \textit{class-specific} keyword extraction with a case study in the German business registry.
    \item We propose a class-specific keyword extraction pipeline that improves upon an existing transformer-based method. Our code is found at \url{https://github.com/sjmeis/CSKE}.
    \item We achieve a new standard for extracting class-specific keywords, measured in a comparative analysis with multiple metrics.
\end{enumerate}

\section{Related Work}
A recent survey structures 167 keyword extraction approaches from the literature \cite{XIE2023103382}. We focus on unsupervised extraction approaches, which can generally be characterized as either statistics-, graph-, or embedding-based, while \textit{TF-IDF} is a common frequency-based baseline method \cite{Papagiannopoulou2020ARO}. 

\textit{YAKE} uses a set of different statistical metrics, including word casing, word position, word frequency, and more, to extract keyphrases from text \cite{CAMPOS2020257}. \textit{TextRank} uses \ac{pos} filters to extract noun phrase candidates that are added to a graph as nodes while adding an edge between nodes if the words co-occur within a defined window \cite{mihalcea-tarau-2004-textrank,Page1999ThePC}. \textit{SingleRank} improves upon the TextRank approach by adding weights to edges based on word co-occurrences \cite{wan-xiao-2008-collabrank}. \textit{RAKE} leverages a word co-occurrence graph and assigns a number of scores to aid in ranking keyword candidates \cite{Rose2010AutomaticKE}. Knowledge Graphs can also be used to incorporate semantics for keyword or keyphrase extraction \cite{Shi2017KeyphraseEU}. \textit{EmbedRank} leverages Doc2Vec \cite{pmlr-v32-le14} and Sent2Vec \cite{pagliardini-etal-2018-unsupervised} embeddings to rank candidate keywords for extraction \cite{bennani-smires-etal-2018-simple}. In a similar way, \textit{PatternRank} uses a combination of sentence embeddings and \textit{POS} filters \cite{kdir22}. Further, Language Model-based approaches have been introduced, for example using BERT \cite{devlin-etal-2019-bert}, for automatic extraction of keywords and keyphrases \cite{sammet-krestel-2023-domain,song-etal-2023-survey}.

\section{A Class-Specific Keyword Extraction Pipeline}
\begin{figure*}
    \centering
    \includegraphics[scale=0.48]{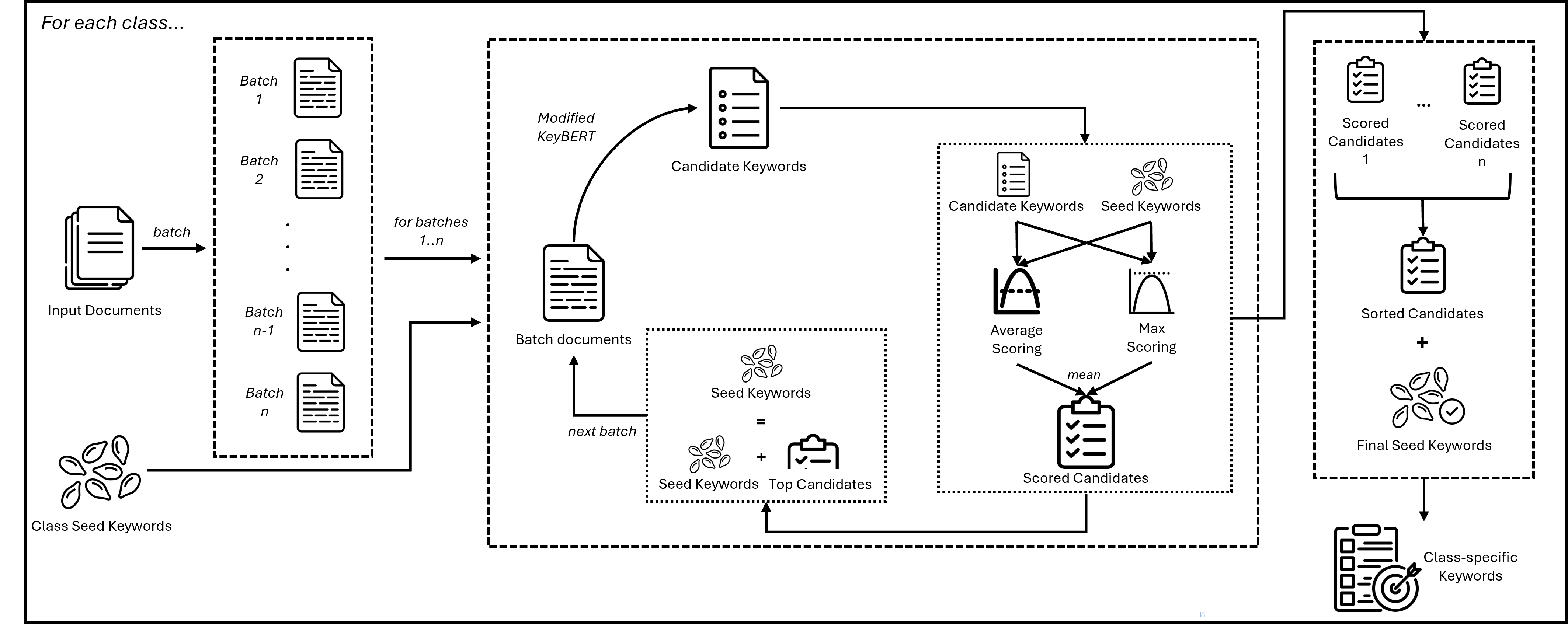}
    \caption{Our class-specific keyword extraction pipeline. With a document corpus and class-specific keyword sets as inputs, we iterate sequentially over batches of the corpus, using a modified \textsc{KeyBERT} and a two-part scoring scheme. Top keywords are added to the seed keywords for the next iteration, until a final set of keywords is achieved.}
    \label{fig:konvens}
\end{figure*}

In this section, we outline in detail our proposed class-specific keyword extraction pipeline. The pipeline is illustrated in Figure \ref{fig:konvens}.

\paragraph{Preliminaries}
For our pipeline, we assume three preliminary requirements:
\begin{enumerate}
    \itemsep -0.3em
    \item \textbf{Document corpus}: unstructured text documents from any domain, from which meaningful information can be extracted.
    \item \textbf{Pre-defined classes}: a set of one or more \textit{classes}, each of which represents a distinct and well-defined concept.
    \item \textbf{Class-specific seed keywords}: for each defined class, a set of \textit{seed keywords} is available. Seed keywords are keywords that are representative of a particular class and can be used as a foundation for guided keyword extraction.
\end{enumerate}

\paragraph{An Iterative Method}
Given a sizeable document corpus, we propose to process the corpus in \textit{batches}, allowing for an iterative method, where each iteration \say{learns} from the previous.

For each iteration (on one batch), the first step is to extract keywords from the batch's documents in a \textit{guided} manner. For this, we modify the popular \textsc{KeyBERT} package, specifically the \textit{guided} functionality. In the current version of \textsc{KeyBERT} the guided functionality by default takes a set of seed keywords as input parameters, and uses a weighted average of seed keyword embeddings and document embeddings to extract candidate keywords. As we place a focus on \textit{class-specific} seed keywords, we make the modification for \textit{KeyBERT} to focus 100\% on the seed keyword embeddings. After this modified version is run on the entire batch, the output is a list of \textit{guided} candidate keywords (i.e., from the seed keywords).

Following the above, we employ a two-part scoring scheme to \say{reorder} the candidates. In particular, we use the following two scores:
\begin{itemize}
    \itemsep -0.2em
    \item \textbf{Average Scoring}: the embedding of each candidate is compared against each seed keyword embedding, using cosine similarity, and these results are averaged for the \textit{average score}.
    \item \textbf{Max Scoring}: similar to \textit{average scoring}, but only the maximum cosine similarity score is kept, resulting in the \textit{max score}.
\end{itemize}
We use the mean of \textit{average score} and \textit{max score} for the final candidate score, and all candidates for a batch are reordered based on this final score. The intuition behind such a scoring scheme is that an ideal keyword is both similar in meaning to one seed keyword, but also generally similar to all seed keywords, suggesting that such a keyword is also representative of the class in question.

The final step within one iteration includes taking the top-scoring candidates and adding them to the set of seed keywords. In doing so, we can iteratively \say{expand} the class-specific seed keywords, thus also expanding the comprehensiveness of these seeds. To do this, we define two parameters: (1) \textit{percentile\_newseed}, defining above which percentile of scores to consider (default: $99$), and (2) \textit{number\_newseed}, defining how many new seed keywords to add per iteration (default: $3$). Thus in the default setting, after each iteration (except the last), a maximum of 3 keywords from the top 99th percentile are added to the set of seed keywords.

\paragraph{Class-specific Keyword Set}
The output of each iteration is a set of scored candidate keywords. After all batches are processed, all scored candidates are merged and sorted. A \textit{topk} parameter governs how many of the keywords to return, with seed keywords always being placed at the top of the list.

\section{Experimental Setup and Results}

\begin{table*}[ht!]
\centering
\resizebox{0.95\textwidth}{!}{
\renewcommand{\arraystretch}{0.8}
\begin{tabular}{ll|cccc||c}
 &  & \multicolumn{1}{c}{Precision@10} & \multicolumn{1}{c}{Precision@25} & \multicolumn{1}{c}{Precision@50} & \multicolumn{1}{c||}{Precision@100} & \multicolumn{1}{c}{Average} \\ \hline
\multicolumn{1}{l|}{} & \textsc{RAKE} & 0.95 & 1.33 & 1.71 & 1.42 & 1.36 \\
\multicolumn{1}{l|}{} & \textsc{YAKE} & 3.33 & 3.24 & 2.38 & 1.81 & 2.69 \\
\multicolumn{1}{c|}{Exact Match} & \textsc{KeyBERT} & 1.90 & 1.71 & 1.71 & 1.05 & 1.60 \\
\multicolumn{1}{l|}{} & Guided \textsc{KeyBERT} & 2.38 & 1.90 & 1.90 & 1.24 & 1.86 \\
\multicolumn{1}{l|}{} & Ours & \textbf{28.10} & \textbf{22.67} & \textbf{13.62} & \textbf{8.33} & \textbf{18.23} \\ \hline
\multicolumn{1}{l|}{} & \textsc{RAKE} & 1.43 & 1.52 & 1.90 & 1.76 & 1.65 \\
\multicolumn{1}{l|}{} & \textsc{YAKE} & 2.38 & 3.24 & 2.67 & 2.33 & 2.65 \\
\multicolumn{1}{c|}{Lemma Match} & \textsc{KeyBERT} & 1.90 & 1.90 & 1.81 & 1.29 & 1.73 \\
\multicolumn{1}{l|}{} & Guided \textsc{KeyBERT} & 2.38 & 1.90 & 2.10 & 1.48 & 1.96 \\
\multicolumn{1}{l|}{} & Ours & \textbf{21.43} & \textbf{20.76} & \textbf{13.43} & \textbf{9.00} & \textbf{16.15} \\ \hline
\multicolumn{1}{l|}{} & \textsc{RAKE} & 62.60 & 61.63 & 61.95 & 59.95 & 61.29 \\
\multicolumn{1}{l|}{} & \textsc{YAKE} & 65.91 & 65.10 & 62.98 & 59.58 & 63.39 \\
\multicolumn{1}{c|}{Fuzzy Match} & \textsc{KeyBERT} & 60.42 & 60.49 & 59.55 & 57.16 & 59.41 \\
\multicolumn{1}{l|}{} & Guided \textsc{KeyBERT} & 60.67 & 60.62 & 59.95 & 57.42 & 59.67 \\
\multicolumn{1}{l|}{} & Ours & \textbf{78.19} & \textbf{75.21} & \textbf{72.93} & \textbf{67.54} & \textbf{73.47} \\ \hline
\multicolumn{1}{l|}{} & \textsc{RAKE} & 77.54 & 79.48 & 79.73 & 79.83 & 79.14 \\
\multicolumn{1}{l|}{} & \textsc{YAKE} & 82.48 & 83.52 & 82.69 & 82.05 & 83.13 \\
\multicolumn{1}{c|}{CS Match} & \textsc{KeyBERT} & 76.73 & 77.39 & 77.09 & 76.86 & 77.02 \\
\multicolumn{1}{l|}{} & Guided \textsc{KeyBERT} & 77.30 & 77.76 & 77.66 & 77.36 & 77.52 \\
\multicolumn{1}{l|}{} & Ours & \textbf{86.32} & \textbf{86.82} & \textbf{86.02} & \textbf{85.36} & \textbf{86.13} \\ \hline \hline
\multicolumn{1}{l|}{} & \textsc{RAKE} & 35.70 & 36.02 & 36.37 & 35.52 & 35.91 \\
\multicolumn{1}{l|}{} & \textsc{YAKE} & 38.90 & 38.69 & 37.62 & 36.40 & 37.90 \\
\multicolumn{1}{c|}{Average Match} & \textsc{KeyBERT} & 34.88 & 35.14 & 34.99 & 34.16 & 34.79 \\
\multicolumn{1}{l|}{} & Guided \textsc{KeyBERT} & 35.21 & 35.31 & 35.38 & 34.44 & 35.08 \\
\multicolumn{1}{l|}{} & Ours & \textbf{53.51} & \textbf{51.41} & \textbf{46.50} & \textbf{42.56} & \textbf{48.49}
\end{tabular}
}
\caption{Class-specific Keyword Extraction Evaluation Results. For each scoring scheme, the highest score for each $k$ is \textbf{bolded}. The average in the right column represents the average of the four evaluated $k$ values. \textit{Average Match} denotes the average score achieved by a method for one $k$ but across all four scores. Examples of extracted keywords for each approach are provided in Appendix \ref{sec:appendix}.}
\label{tab:results}
\end{table*}

Our experimental setup aims to evaluate the ability of our proposed method to extract class-specific keywords, in comparison to previous approaches. As opposed to typical keyword extraction evaluations, our evaluation tests the ability of a method to extract a set of class-specific keywords from a corpus, rather than generic keywords from documents.

\paragraph{Dataset}
We use a dataset of the German business registry (\textit{Deutsches Handelsregister}) records, which contains 2.37 million business purpose records structured by Fusionbase\footnote{\url{https://fusionbase.com}}. 
The goal is to classify each business into an economic sector, according to the scheme proposed by the German Ministry of Statistics (\textit{Statistiches Bundesamt}), called the \textit{WZ 2008} (\textit{Klassifikation der Wirtschaftszweige, Ausgabe 2008})\footnote{\label{wz}\scriptsize\url{https://www.destatis.de/DE/Methoden/Klassifikationen/Gueter-Wirtschaftsklassifikationen/klassifikation-wz-2008.html}}.
In this work, we model the evaluation on the above dataset as a class-specific keyword extraction task, where the goal is to extract meaningful keywords for each of the 21 top-level economic sectors in the WZ 2008. For evaluation purposes, we use a random sample of 10,000 rows from the larger dataset\footnote{This sample can be found in our code repository.}.

It should be noted that we only investigate the extraction of unigram keywords. For the extraction of German keywords, this is advantageous due to the relatively high frequency of nominal compounds in the German language. Thus, meaningful keywords can be extracted in an efficient manner. However, this comes with two limitations: (1) not all \textit{keyphrases} will be caught, thus sometimes leading to incomplete keywords (see \say{Dicke} in Listing \ref{lst:seed1}, which means \textit{thick} translated to English), and (2) the results achieved for German language datasets may not be directly generalizable to English.

\paragraph{Keyword Extraction Methods}
For a comparative analysis, we test our method against four methods: (1) \textsc{RAKE} \cite{Rose2010AutomaticKE}, (2) \textsc{YAKE} \cite{CAMPOS2020257}, (3) \textsc{KeyBERT}, and (4) Guided \textsc{KeyBERT}. Note that \textsc{RAKE} and \textsc{YAKE} do not offer any mechanism for guided keyword extraction, and thus the resulting keywords are the same for each class. We test our proposed method with the parameter \textit{n\_iterations} (number of batches) set to $5$. \textit{Guided} \textsc{KeyBERT} refers to the use of the optional \texttt{seed\_keywords} parameter, which serves as a direct comparison point to our proposed method (denoted \textit{ours}). For \textsc{KeyBERT} and our method, we use the \textsc{deutsche-telekom/gbert-large-paraphrase-cosine} language model. Note that for comparability, \textsc{KeyBERT} was set only to extract unigram keywords.

\paragraph{Seed Keywords}
For the selection of seed keywords, specifically for Guided \textsc{KeyBERT} and our method, we utilize an existing collection of keywords (\textit{Stichwörter}) provided by the creators of the WZ 2008$^{\ref{wz}}$. As we aim only to extract unigrams, we truncate all keyphrases to the first word if they are longer than one word. From this gold set, we randomly select 10 keywords from each class to serve as the seeds for that class. The rest of the gold set is then used for evaluation. The seed keywords from two classes are presented in Listings \ref{lst:seed1} and \ref{lst:seed2}.
\vspace{3pt}

\scriptsize
\begin{lstlisting}[language=json,firstnumber=1,caption={Seed Keywords for Class A: \textit{Land- und Forstwirtschaft, Fischerei}. Seed keywords marked with an asterisk (*) denote those found in our dataset sample.},captionpos=b,label=lst:seed1]
['Schweinehaltung', 'Holztaxierung',
 'Austernzucht', 'Teichwirtschaft',
 'Tabak'*, 'Dicke',
 'Fischerei'*, 'Seidenraupenzucht',
 'Wild', 'Kassava']
\end{lstlisting}

\begin{lstlisting}[language=json,firstnumber=1,caption={Seed Keywords for Class D: \textit{Energieversorgung}. Seed keywords marked with an asterisk (*) denote those found in our dataset sample.},captionpos=b,label=lst:seed2]
['Heizkraftwerke'*, 'Elektrizitaetserzeugung',
 'Blockheizkraftwerk'*, 'Waermeversorgung',
 'Solarstromerzeugung', 'Bereitstellung'*,
 'Energieversorgung'*, 'Windparks'*,
 'Spaltgaserzeugung', 'Kokereigasgewinnung']
\end{lstlisting}

\normalsize
\paragraph{Metrics}
With the keywords sets from each of the tested methods, we evaluate the accuracy of the keywords on two dimensions: (1) \textit{precision@K}, where the number of correct keywords amongst the top $K$ output keywords is counted, and (2) \textit{matching method}, where the meaning of \say{correct} is varied. For $K$, we choose $K \in \{10,25,50,100\}$, and for matching method, we use four approaches:
\begin{itemize}
    \itemsep -0.4em
    \item \textbf{Exact string match}: a correct keyword is counted if the extracted keyword is found \textit{exactly} in the gold set of keywords.
    \item \textbf{Lemma match}: a correct keyword is counted if the \textit{lemmatized} version of the keyword is found in the \textit{lemmatized} gold set of keywords \cite{zesch-gurevych-2009-approximate}.
    \item \textbf{Fuzzy string match}: the \say{correctness} of a keyword is not binary, but rather is represented by the closest fuzzy string match score, using the Python package \textsc{thefuzz}.
    \item \textbf{Cosine similarity match}: the correctness of a keyword is measured by its highest cosine similarity to any of the gold keywords.
\end{itemize}
For cosine similarity, the \textsc{deepset/gbert-base} model is used, so as not to use the same base model used with the keyword extraction process.

\paragraph{Results}

Table \ref{tab:results} presents the results of the above-described experiments. Note that for the evaluation of extracted keywords against the gold set, we only include keywords in the gold set that appear (in lemmatized form for \textit{lemma match}) in the 10k sample of the German business registry data.

We can observe that our approach outperforms all other methods in \textit{class-specific} keyword extraction. The performance of our approach is particularly strong in the exact match and lemma match evaluations, indicating it is well suited to extract class-specific gold keywords as defined by the creators of the WZ 2008$^{\ref{wz}}$ classification scheme. Notably, even the Guided \textsc{KeyBERT} method, designed to extract keywords similar to provided seed keywords, performs significantly worse than our approach. Looking to the results, we see that the guided version of \textsc{KeyBERT} often only shows improvements over the base version when more extracted results are considered. This implies that while some class-specific keywords are found, they are not ranked as high as other keywords. Ultimately, we conclude that our approach achieves state-of-the-art results for \textit{class-specific} keyword extraction, a point that is supported by a qualitative analysis of example outputs in Appendix \ref{sec:appendix}.

\section{Conclusion}
\label{sec:conclusion}
We present a class-specific keyword extraction pipeline which outperforms previous methods in identifying keywords related to a predefined \textit{class}. Our evaluation results exhibit the strong performance of our method in the task of retrieving keywords specific to particular German economic sectors. These results make a compelling case for the continued study of class-specific keyword extraction as an improvement to non-guided approaches.

As points for future work, we propose more rigorous evaluation of our method from two perspectives: (1) an ablation study on the effect of the \textit{n\_iterations}, \textit{number\_newseed}, \textit{percentile\_newseed}, and \textit{topk} parameters, in particular to study their relevance for class-specific keyword extraction, and (2) evaluation of our method beyond the German language, firstly with English.

\section*{Acknowledgments}
This work has been supported by the BayVFP Digitalization grant DIK-2210-0028//D1K0475102 (CreateData4AI) from the Bavarian Ministry of Economic Affairs, Regional Development and Energy. The project is performed in collaboration with Fusionbase GmbH, whom we thank for the Business Registry data access and for the guidance.

\section*{Limitations}
The primary limitation of our work is the lack of evaluation of the various parameters of our method, as discussed in Section \ref{sec:conclusion}. Evaluating a range of values would strengthen the work in determining the individual effect of each parameter.

The second limitation involves the relatively limited scope both in domain and language. In particular, we focus our case study only on the German Business Registry, and we do not generalize beyond this to different domains or languages.

\section*{Ethics Statement}
An ethical consideration comes with the use of the German Business Registry dataset, which is directly tied to real-world businesses, potentially raising privacy concerns. However, this is mitigated by the fact that the data is public and business owners are aware of this when drafting their entries.

\bibliography{anthology,custom}

\appendix
\section{Extracted Keyword Examples}
\label{sec:appendix}
\footnotesize

\begin{lstlisting}[language=json,firstnumber=1,caption={Sample extracted keywords for Class A, from the 10:25 top keywords for each method.},captionpos=b]
{'rake': ['analyse',
  'entwicklung',
  'software',
  'programmen',
  'weiterentwicklung',
  'verkauf',
  'vermietung',
  'domainadressen',
  'housing',
  'domainverwaltung',
  'peering',
  'administration',
  'saemtliche',
  'handel',
  'insbesondere'],
'yake': ['uebernahme',
 'dienstleistungen',
 'geschaefte',
 'beteiligung',
 'verkauf',
 'entwicklung',
 'vermittlung',
 'geschaeftsfuehrung',
 'beratung',
 'herstellung',
 'beteiligungen',
 'taetigkeiten',
 'erbringung',
 'bereich',
 'immobilien'],
'keybert': ['landschaftsbau',
  'photovoltaik',
  'elektroinstallationen',
  'masskleidung',
  'landschaftsmusikfestivals',
  'systemgastronomie',
  'bauleistungen',
  'reisebueros',
  'immobilien',
  'physiotherapie',
  'wasserinstallationsarbeiten',
  'diskothek',
  'nassbaggerarbeiten',
  'druckereierzeugnissen',
  'zahntechnischen'],
'guided_keybert': ['landschaftsbau',
  'elektroinstallationen',
  'photovoltaik',
  'systemgastronomie',
  'landschaftsmusikfestivals',
  'masskleidung',
  'bauleistungen',
  'reisebueros',
  'immobilien',
  'diskothek',
  'wasserinstallationsarbeiten',
  'druckereierzeugnissen',
  'nassbaggerarbeiten',
  'physiotherapie',
  'zahntechnischen'],
'ours': ['zucht',
  'fuger',
  'getreide',
  'spenglerei',
  'verpachtungen',
  'veraeu',
  'frachten',
  'fracht',
  'schalungen',
  'verpachtung',
  'beund',
  'kalk',
  'schalung',
  'holzwaren',
  'haefte']
}
\end{lstlisting}

\begin{lstlisting}[language=json,firstnumber=1,caption={Sample extracted keywords for Class D, from the 10:25 top keywords for each method.},captionpos=b]
{'rake': ['analyse',
  'entwicklung',
  'software',
  'programmen',
  'weiterentwicklung',
  'verkauf',
  'vermietung',
  'domainadressen',
  'housing',
  'domainverwaltung',
  'peering',
  'administration',
  'saemtliche',
  'handel',
  'insbesondere'],
'yake': ['uebernahme',
 'dienstleistungen',
 'geschaefte',
 'beteiligung',
 'verkauf',
 'entwicklung',
 'vermittlung',
 'geschaeftsfuehrung',
 'beratung',
 'herstellung',
 'beteiligungen',
 'taetigkeiten',
 'erbringung',
 'bereich',
 'immobilien'],
'keybert': ['immobilien',
  'delaware',
  'verkauf',
  'pizzalieferservices',
  'unternehmens',
  'ambulanten',
  'eingliederungshilfe',
  'gesellschaftsbeteiligungen',
  'bebauung',
  'schulverwaltungssoftware',
  'geschaeftsfuehrung',
  'textilzubehoer',
  'masskleidung',
  'motorradzubehoerteilen',
  'casinobetriebe'],
'guided_keybert': ['immobilien',
  'delaware',
  'kraftfahrzeugen',
  'pizzalieferservices',
  'unternehmens',
  'ambulanten',
  'eingliederungshilfe',
  'gesellschaftsbeteiligungen',
  'bebauung',
  'schulverwaltungssoftware',
  'geschaeftsfuehrung',
  'textilzubehoer',
  'masskleidung',
  'motorradzubehoerteilen',
  'casinobetriebe'],
'ours': ['energieanlagen',
  'energieerzeugungsanlagen',
  'energieerzeugung',
  'energietechnik',
  'energieversorgungs',
  'energietechnischen',
  'energieprodukten',
  'stromerzeugungsanlagen',
  'energiegewinnung',
  'energietraeger',
  'energietraegern',
  'energiequellen',
  'ernergieanlagen',
  'energie',
  'stromerzeugern']}
\end{lstlisting}

\end{document}